\documentclass[letterpaper, 10 pt, conference]{ieeeconf}  % Comment this line out if you need a4paper

\IEEEoverridecommandlockouts                              % This command is only needed if 
\usepackage{amsmath,amsfonts,bm}

\def\eqref#1{equation~\ref{#1}}
\def\1{\bm{1}}

\DeclareMathAlphabet{\mathsfit}{\encodingdefault}{\sfdefault}{m}{sl}
\SetMathAlphabet{\mathsfit}{bold}{\encodingdefault}{\sfdefault}{bx}{n}

\newcommand{\E}{\mathbb{E}}

\newcommand{\KL}{D_{\mathrm{KL}}}

\usepackage[english]{babel}

\usepackage{caption}
\usepackage{subcaption}
\usepackage{graphicx}
\usepackage{booktabs}
\usepackage{stfloats}
\usepackage{url}

\title{\LARGE \bf
Learning to Navigate from Scratch using World Models\\ and Curiosity: the Good, the Bad, and the Ugly
}

\author{Daria de Tinguy\textsuperscript{*}\textsuperscript{1}, Sven Remmery\textsuperscript{*}\textsuperscript{1}, Pietro Mazzaglia\textsuperscript{1}, Tim Verbelen\textsuperscript{2} and Bart Dhoedt\textsuperscript{1}% <-this % stops a space
\thanks{*Equal contribution}% <-this % stops a space
\thanks{\textsuperscript{1}Department of Information Technology, University of Ghent, Ghent, Belgium,\textsuperscript{2}Verses AI, Vancouver, Canada
        {\tt\small \{Correspondence to: Daria de Tinguy <Daria.detinguy at
ugent.be>}}%
}
\begin{document}

\maketitle
\thispagestyle{empty}
\pagestyle{empty}

%- Section I: The meaning of ""mitigated results"" is unclear. --> replaced by "resulting in a limited exploration"
%- Section II: Differences between previous research and your approach can make it easier to understand where the paper stands.
%- Some sentences might be modified because citation numbers are not included as words. For example, these might be avoided: ""The approach described in [8] proposes ..."" and ""[9] centres on the maximisation of ...""  -> DONE
%- Which specific model of GPU did you use? -> NO IDEA!!! Did Sven have cluster 9? 

%- What does it mean that the observation of future t+1 is obtained from the environment? When the current time is t, x_t+1 should not yet be observed. --> I added: " it moves from time $t$ to $t+1$, therefore receiving a new observation $x_{t+1}$"
%- Does the expected reward refer to r_expl? The relationship between equations 2 and 3 is unclear.
%- Are maps acquired by SLAM used for self-localization and navigation in ROS? -> I don't get the question at all (how did he get to that?)? I added a sentence explaining what ROS is... about SLAM it is already defined in related work. Don't know what else to do.

%- Several reference entries are missing." --> !!! I don't see any missing ref. 

%%%%%%%%%%%%%%%%%%%%%%%%%%%%%%%%%%%%%%%%%%%%%%%%%%%%%%%%%%%%%%%%%%%%%%%%%%%%%%%%
\begin{abstract}

%This electronic document is a IEEE template. The various components of your paper [title, text, heads, etc.] are already defined on the style sheet, as illustrated by the portions given in this document.

Learning to navigate unknown environments from scratch is a challenging problem. This work presents a system that integrates world models with curiosity-driven exploration for autonomous navigation in new environments. We evaluate performance through simulations and real-world experiments of varying scales and complexities. In simulated environments, the approach rapidly and comprehensively explores the surroundings. Real-world scenarios introduce additional challenges. Despite demonstrating promise in a small controlled environment, we acknowledge that larger and dynamic environments can pose challenges for the current system. Our analysis emphasizes the significance of developing adaptable and robust world models that can handle environmental changes to prevent repetitive exploration of the same areas.
% Our findings highlight the importance of scalability and robustness in real-world navigation systems as well as the integration of world models, curiosity-driven exploration, and reinforcement learning in real world scenarios.
\end{abstract}

%%%%%%%%%%%%%%%%%%%%%%%%%%%%%%%%%%%%%%%%%%%%%%%%%%%%%%%%%%%%%%%%%%%%%%%%%%%%%%%%
\section{Introduction}

Navigation is an extensively studied domain within the field of robotics, where classical solutions typically combine mapping and planning routines. Traditional approaches involve space representation in grid \cite{geo_map,2Dmap} and/or topological maps \cite{sp2atm,grid_topo} based on observed data, over which motion planning algorithms are employed to guide robots through the environment. However, the efficacy of these approaches is inevitably intertwined with the critical hurdle of exploration.

Exploration, as a crucial part of navigation, is a puzzle. Uncovering new regions, detecting obstacles, and deducing viable approaches, all while learning a sensible model of the environment, are all part of the process.  In light of these challenges, recent research has shed light on the integration of machine learning techniques to augment the autonomy and adaptability of models, enabling them to adeptly navigate uncharted scenarios \cite{robot_nav_0,exp_learning_survey}. 

% Specifically, -> removed as it doesn't connect with previous paragraph about exploration
The combination of world models with reinforcement learning (RL) \cite{Dreamerv3,imitation_learning_AIF,RL-Visual_Nav,RECON} has emerged as a powerful data-driven strategy for enhancing decision-making prowess within real-world contexts \cite{RL_GM_intel_agent_H_brain,RL_GM_benefit_proof_3D_env}. However, it is worth noting that the application of this approach to real-world scenarios has predominantly centered around constrained workspaces, such as manipulation tasks \cite{RL_GM_manipulation,RL_GM_manipulation2,RL_GM_manipulation3,RL_GM_manipulation4,Ferraro}, or locomotion endeavors \cite{muscle_learning1,muscle_learning2,muscle_learning3}.

This work proposes and evaluates a machine learning system for learning robotic navigation from scratch, using world models, for learning a representation of the environment, model-based RL, to drive low-level actions, and curiosity-driven exploration, to discover reachable areas in the environment. We aim to analyze whether current trends in machine learning, such as world models and unsupervised RL, already constitute valuable means for real-world robotic navigation, from simulation to real-world applications of diverse scale and complexity. 

\textbf{Contributions.} Our work can be summarized as follows:
\begin{itemize}
\item We present a system that combines world models for learning actions \cite{Hafner2022DreamerV2} with curiosity-driven exploration \cite{Mazzaglia2022LBS} (Section \ref{sec:method}).
% which is adequate to the navigation problem as it addresses the POMDP difficulties through a memory-based model -> we can discuss this alteration
\item We evaluate the system in simulation using the Habitat navigation environment \cite{habitat}. Our method outperforms previous state-of-the-art approaches \cite{cbet} when comparing the percentage of the environment explored (\textit{the good}, Section \ref{subsec:simulation}).
\item We design a remote-computing pipeline to apply our system in real-world scenarios and evaluate it in two environments: a controlled small-scale environment, where the agent performs well, and an uncontrolled experiment % (learning in the wild)
in a larger-scale environment, where the agent's exploration capabilities are impaired (\textit{the bad}, Section \ref{subsec:real}).
\item We quantitatively and qualitatively assess our work, demonstrating how the generative world model underperforms when visual aspects of the environment change during training (e.g. objects are rearranged in a room). We observe that the agent's imagined trajectories struggle to match the dynamism of reality (\textit{the ugly}, Section \ref{subsec:analysis}).
\end{itemize}
Overall, our system demonstrates high autonomous exploration performance in simulated and small-scale real-world environments. However, in a larger environment, the agent's performance is hindered by hardware and model limitations, resulting in a limited exploration. We hope that this investigation will drive further developments in this promising research direction.
    
    % This paper presents a generative model trained end-to-end through reinforcement learning in simulated and real world environments ///++ short description model to write once method done -to avoid discrepancies- //. 
    
    % //last part is a bit untasty, what more do we bring on the table. to reformulate//
    % // no mention of POMDP, could be added if we want the intro longer (+ more decsription of models)
    %Submission to CoRL 2023 will be entirely electronic, via a web site (not email). Information about the submission process and \LaTeX{} templates are available on the conference web site at \url{https://corl2023.org/}. For camera ready submission, use the \texttt{final} option for the \texttt{\textbackslash usepackage} command. 

%===============================================================================

\section{Related Work}

\textbf{Robot navigation.} Robotic navigation has been a longstanding challenge, often involving simultaneous localization and mapping (SLAM) techniques. Traditional methods typically focus on building metric (grid) \cite{geo_map,2Dmap}, topological \cite{sp2atm} or both maps \cite{grid_topo} of the environment and then use motion planning algorithms to navigate through it. Despite the progress made, it is acknowledged that these approaches lack autonomy in complex environments \cite{survey_slam}. For example, active SLAM struggles with challenges like predicting robot localization uncertainty, creating semantic-like mappings, and reasoning in dynamic and changing spaces \cite{robot_nav_0,exp_learning_survey}.
    %Those traditional approaches often leave a range of situations to handle case per case, resulting in complex systems . 

Recent research has sought to enhance robotic navigation with machine learning techniques, aiming to imbue models with adaptability and autonomy in new scenarios. Different approaches have been explored to address navigational challenges by acquiring navigational skills from simulation \cite{DL_simu_train1,DL_simu_train2}, human-provided labels \cite{label1,label2}, human demonstration \cite{H_demo} or through lifelong learning \cite{cbet,map_visual_exp_L,end_end_exp_L_SLAM}.

\textbf{World models.} World models have gained traction alongside reinforcement learning (RL) approaches, as they provide the agent with an internal representation of the environment. These models allow agents to simulate and plan future trajectories, thereby supporting decision-making and adaptability \cite{WM}. Combining world models with RL for decision-making enables solving tasks driven by a reward function \cite{Dreamerv3, Hafner2022DreamerV2} and has shown benefits in terms of data-efficiency for real-world problems as well  ~\cite{RL_GM_intel_agent_H_brain,RL_GM_benefit_proof_3D_env, wu2022daydreamer}.
However, notable works combining RL and world models in real-world environments have predominantly focused on specific contexts such as arm manipulation \cite{RL_GM_manipulation,RL_GM_manipulation2,RL_GM_manipulation3,RL_GM_manipulation4,Ferraro} or locomotion control \cite{muscle_learning1,muscle_learning2,muscle_learning3}.

\textbf{Exploration.} 
Extensive literature exists in the context of unsupervised exploration and unsupervised objectives for control \cite{Mazzaglia2022LBS, Pathak2017ICM, yarats2021proto, rnd, houthooft2017vime, Achiam2017SurpriseDeepRL, cbet}, with some of these works also combining exploration with world models applied in simulated manipulation and locomotion environments~\cite{Sekar2020Plan2Explore, Rajeswar2023MasterURLB}. Only a few studies combine world models and RL for navigation, and even fewer propose an exploration behavior not relying on a goal objective to stimulate exploration. The approach described in Nozari \& al.~\cite{imitation_learning_AIF} proposes a generative model aiming to enhance both exploratory and exploitative navigation in unexpected situations without necessarily relying on goal objectives. 
% This work is particularly relevant in situations where traditional RL approaches fail due to an overreliance on minimizing surprise during decision-making. 
Wu \& al. work ~\cite{RL-Visual_Nav} centers on the maximization of mutual information between actions and subsequent observations to facilitate objective-driven navigation, in order to reach an objective. It's worth noting, however, that both of these approaches necessitate expert data for training. 
% The former employs imitation learning, potentially restricting its performance to that of the expert, while the latter relies on pre-conceived topological maps. 
The RECON system \cite{RECON} introduces a strategy enabling exploration and navigation in open-world environments. It is achieved through the utilization of learned goal-conditioned distance models and latent variables that represent visual objectives. The approach fosters goal-directed exploration and is fine-tuned via real-world trial and error. 

% However, this method lacks explicit consideration of the value of information (its usefulness), which could potentially enhance the efficacy of the policy aimed at achieving goals.

%===============================================================================

\section{Method}
\label{sec:method}

Our approach combines two popular trends in reinforcement learning and aims to study them for robotic navigation: world models, and unsupervised exploration.

\textbf{Setting.} 
The control setting for navigation can be formalized as a Partially Observable Markov Decision Process (POMDP) operating at a discrete frequency, for which we use subscripts to indicate the timesteps. In the POMDP, observations $x_t$ are generated at each timestep, based on the internal hidden state of the environment. Actions $a_t$ enable interaction between the agent and the environment. In this work, we use RGB images as observations while actions are the robot's discrete spatial displacements, with respect to its position.

\subsection{World model}
To address the challenges posed by the POMDP, where the agent cannot observe the hidden state of the environment, we adopt a world model using a recurrent state-space model (RSSM) \cite{Hafner2019RSSM} for the dynamics. This allows learning an internal representation of model states $s_t$ that summarizes and reflects the environment dynamics, integrating knowledge over sequences of observations and actions, thanks to the recurrent memory module.

The world model is composed of the following components:
\begin{equation*}
\begin{split}
    \textrm{Posterior:} & \quad q_\phi(s_t|s_{t-1}, a_{t-1}, x_t), \\
    \textrm{Dynamics:} & \quad p_\phi(s_t|s_{t-1}, a_{t-1}), \\
    \textrm{Decoder:} & \quad p_\phi(x_t|s_t). 
\end{split}
\end{equation*}
The model states $s_t$ have both a deterministic component, modeled using the RSSM \cite{Hafner2019RSSM}, and a discrete stochastic component \cite{Hafner2022DreamerV2}, capturing the uncertainty of the dynamics. The remaining modules are CNNs \cite{LeCun1999CNN}, given the high-dimensionality (pixel-based observations) of the input. The model is trained end-to-end by optimizing the evidence lower bound (ELBO) on the log-likelihood of data:
\begin{equation}
\begin{split}
    \mathcal{L}_\textrm{wm} &= \KL [q_\phi(s_t|s_{t-1}, a_{t-1}, x_t) \Vert p_\phi(s_t|s_{t-1}, a_{t-1})] \\ 
    & - \E_{q_\phi(s_t)}[\log p_\phi(x_t|s_t)],
    \label{eq:wmloss}
\end{split}
\end{equation}
where sequences of observations $x_t$ and actions $a_t$ are sampled from a replay buffer, that is collected online by the agent through exploration.

% \textbf{Imagination through RL}. 
Leveraging this generative world model, the agent can engage in simulated scenarios to explore potential actions and their consequences. This imaginative environment allows the agent to learn and refine its decision-making policies through interactions without directly impacting the real-world environment. Combined with RL techniques \cite{Hafner2022DreamerV2, Dreamerv3}, this mechanism enables data-efficient learning of behavior policies ``in imagination".

% The agent utilizes the world model to simulate the consequences of taking specific actions and evaluates their quality by assigning reward values to the simulated outcomes. These reward signals guide the policy network's learning process, encouraging the agent to prioritize actions that lead to favorable outcomes.

% By having this process occurring within the imaginative realm, any unintended consequences is prevented from manifesting in the real-world environment. This imaginative exploration through RL embodies a fundamental principle of our approach, where the agent learns to navigate and adapt within its environment by leveraging the generative world model's insights.

% Let's complete that while the world model retains memory of past states within an episode, the behavior model does not require memory of past experiences, i.e. the policy is Markovian as well.

\subsection{Exploration}

For exploring the space to navigate, we are interested in collecting data that is informative about new areas of the environment. 
Inspired by curiosity-driven approaches \cite{Mazzaglia2022LBS, houthooft2017vime, Pathak2017ICM}, we would like to measure the amount of information that is gained by the model when facing a new environment's transition and use that as an intrinsic reward to foster exploration in RL, using the world model. Every time the agent takes action $a_t$ while being in the internal state $s_t$, it moves from time $t$ to $t+1$, therefore receiving a new observation $x_{t+1}$ from the environment, that will make the internal state transition to $s_{t+1}$. 

Following \cite{Mazzaglia2022LBS}, the information gained by the world model with the transition can be approximated using the KL divergence between the latent posterior and dynamics and adopted as an intrinsic reward for RL as follows:
\begin{equation}
\begin{split}
    r_\textrm{expl} &= \mathcal{I}(s_{t+1};x_{t+1}|s_t, a_t) \\
    &\approx \KL [q_\theta(s_{t+1}|s_t, a_t, x_{t+1}) \Vert p_\theta(s_{t+1}|s_t, a_t) ]
\end{split}
\label{eq:expl_ac}
\end{equation}
% that is the KL divergence between our approximate posterior model $q(s_{t+1})$, which has access to the privileged information in $o_{t+1}$, and the latent dynamics prior, $p(s_{t+1})$. 
The above term can be efficiently computed by comparing the distributions predicted by the latent dynamics and the latent posterior components. The signal provided should encourage the agent to collect transitions where the predictions are more uncertain or erroneous. %, compared to the true (app

% Assuming the LBS posterior approximates the true posterior of the system dynamics, the cross-entropy term closely resembles the `surprisal' bonus adopted in other works \cite{Achiam2017SurpriseDeepRL, Pathak2017ICM, Burda2019LSCuriosity}.  
% Using LBS can then be seen as maximizing the `surprisal', while trying to avoid high-entropy, stochastic states.

% we shape rewards following LBS \cite{Mazzaglia2021LBS}, which has proven to be a reliable exploration method when using world models \cite{Rajeswar2022MBPT}. 

% Exploration rewards approximate the information gain between observations and model states, computed by measuring the KL divergence between the posterior and the prior of the model.

In order to leverage the world model for learning actions, we train the exploration actor and critic networks in imagination \cite{Hafner2022DreamerV2}, to maximize the expected reward presented in Equation \ref{eq:expl_ac}.  Actor and critic networks are implemented as MLP and defined as follows:
\begin{equation}
\begin{split}
        \textrm{Expl.\ actor:} & \quad \pi_\textrm{expl}(a_t|s_t), \\
        \textrm{Expl.\ critic:} & \quad v_\textrm{expl}(s_t),
\end{split}
\label{eq:expl_ac2}
\end{equation}

Please refer to the Appendix \ref{app:hyperparameters} for the hyperparameters of the world model and the exploration actor-critic.

% \begin{equation*}
%         r_\textrm{expl} = \KL[ q_\phi(s_t|s_{t-1}, a_{t-1}, x_t) \Vert p_\phi(s_t|s_{t-1}, a_{t-1})],
% \end{equation*}

\subsection{Lifelong learning}

All agents are trained in an episodic setting. In each episode, the agent interacts $N$ steps with the environment, storing its actions and observations in a replay buffer. Next, we train the model for $K$ iterations on the replay buffer data and update the agent's world model and policies with the new parameters. Finally, the agent is respawned at the starting position, and a new episode starts.

In the real world, we cannot respawn the agent automatically. In this case, the starting position is defined as the charging station, and we command the robot to return to the charging station at the end of the episode. To enable this, we set up a navigation stack that fuses odometry, IMU and LiDAR data through an Extended Kalman Filter (EKF) \cite{EKF} enabling the robot to be guided back to the charging station. More information about the auto docking strategy can be found in Appendix \ref{app:dock}. Each time the agent docks, the latest episode is sent to a local cloud storage and we execute the model train iterations on a GPU server, using a GTX 1080. 

In our experiments, we used $N = 500$ steps as episode length, and $K= 100$ for updating the models between episodes, sampling sequences of steps over all the collected episodes. % to prevent catastrophic forgetting.

\begin{figure*}[htb!]
     \centering
     \begin{subfigure}[b]{0.21\textwidth}
         %\centering
         \includegraphics[width=\textwidth]{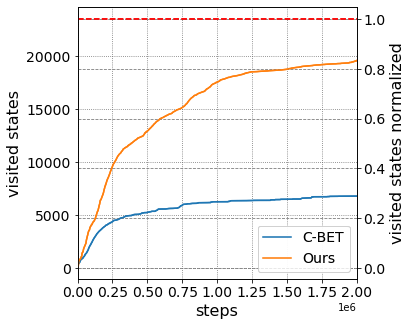}
         \caption{Apartment 0}
         \label{fig:apartment 0}
     \end{subfigure}
     \hfill
     \begin{subfigure}[b]{0.21\textwidth}
         %\centering
         \includegraphics[width=\textwidth]{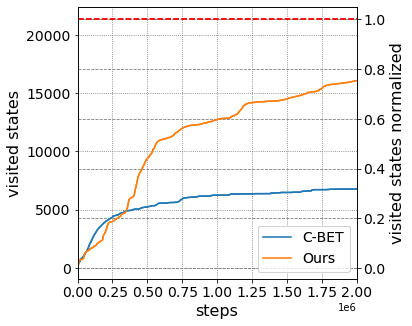}
         \caption{Apartment 1}
         \label{fig:apartment 1}
     \end{subfigure}
     \hfill
     \begin{subfigure}[b]{0.21\textwidth}
         %\centering
         \includegraphics[width=\textwidth]{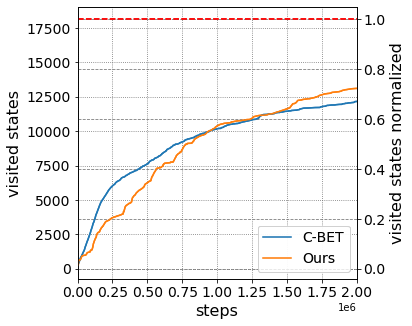}
         \caption{Apartment 2}
         \label{fig:apartment 2}
     \end{subfigure}
     \hfill
     \begin{subfigure}[b]{0.3\textwidth}
         %\centering
         \includegraphics[width=\textwidth]{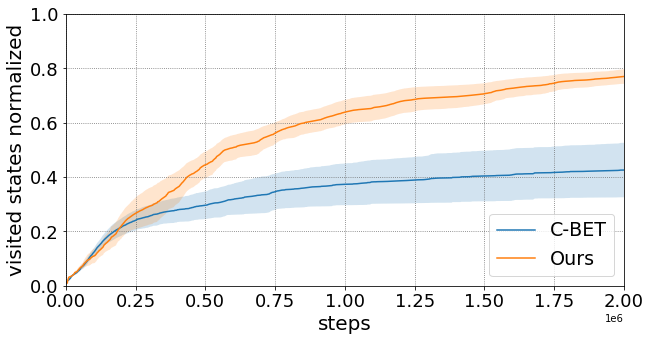}
         \caption{Over all runs}
         \label{fig:all_apps}
     \end{subfigure}
        \caption{C-BET and our agent's performnace comparison over 3 Habitat environments. For each figure, on the left, the number of visitable states the agents reached, the dotted line representing the environment maximum visitable states. On the right, the coverage percentage the agent reached. (d) shows the mean of the normalized visited states over all the environments.}
        \label{fig:three appartments}
\end{figure*}

\begin{figure*}[t!]
     \centering
     \hfill
     \begin{subfigure}[b]{0.3\textwidth}
         \centering
         \includegraphics[width=\textwidth]{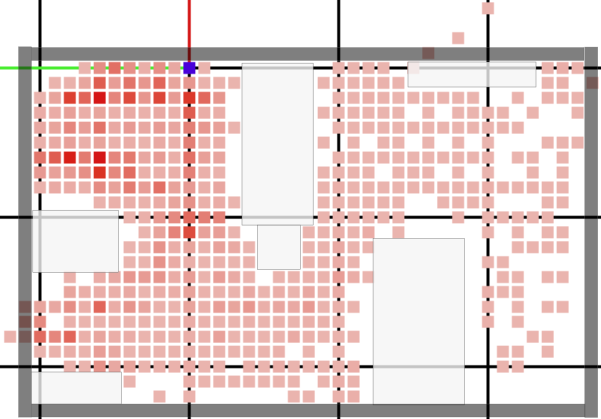}
         \caption{Coverage heatmap.}
         \label{fig:heatmap_garage}
     \end{subfigure}
     \hfill
     \begin{subfigure}[b]{0.26\textwidth}
         \centering
         \includegraphics[width=\textwidth]{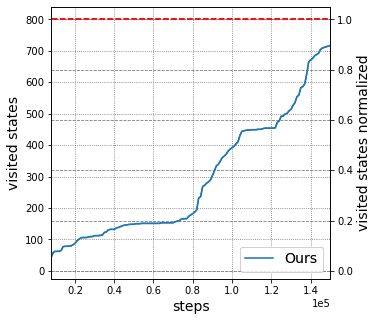}
         \caption{Run 1}
         \label{fig:run 1}
     \end{subfigure}
     \hfill
     \begin{subfigure}[b]{0.26\textwidth}
         \centering
         \includegraphics[width=\textwidth]{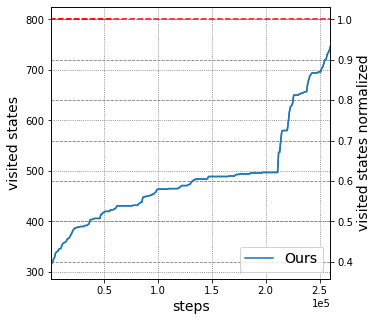}
         \caption{Run 2}
         \label{fig:run 2}
     \end{subfigure}
     \hfill
    \caption{(a) Coverage heatmap of the environment with a coarse positioning of the obstacles. Blue square is the starting position. We can see that Run 1 (b) outperforms Run 2 (c), as in the latter the agent discovers the furthest area only after 250k steps. The blue curve shows the number of visited states covered over the number of total steps, the dotted red line displaying the maximum number of states the environment.}
    \label{fig:real_small}
\end{figure*}

\begin{figure*}[b!]
     \hfill
     \begin{subfigure}[b]{0.4\textwidth}
         \centering
    \includegraphics[width=\textwidth]{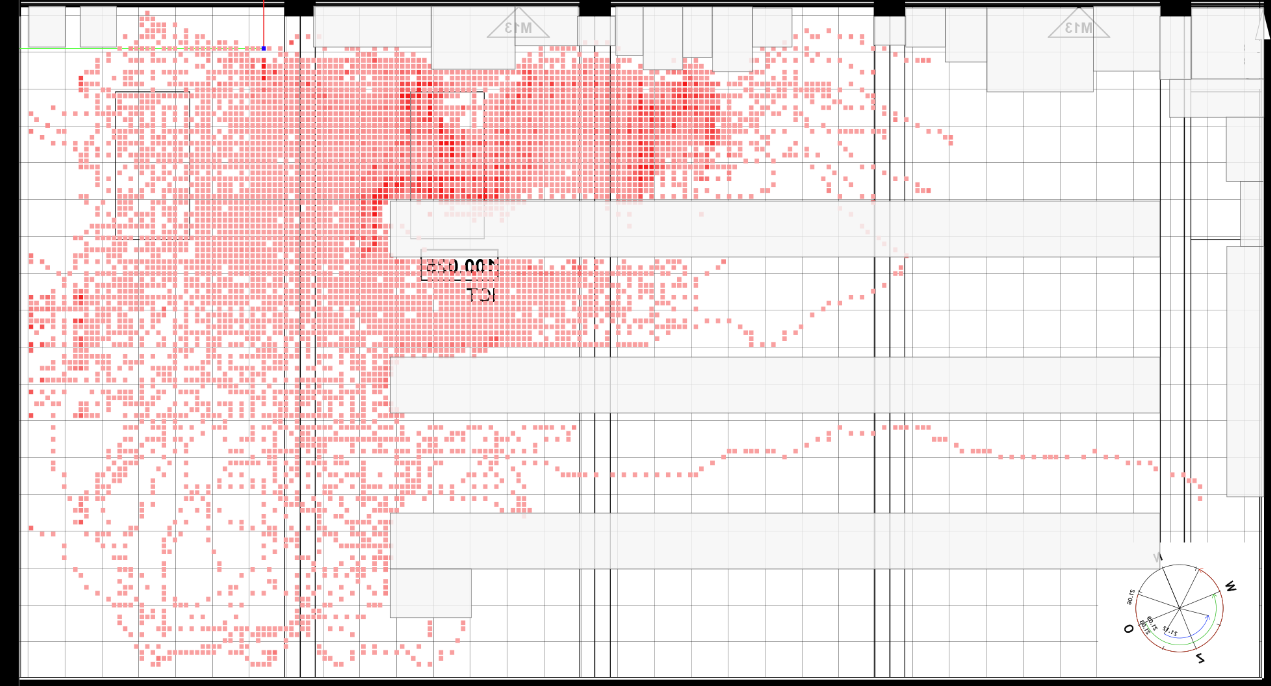}
    \caption{Heatmap of the environment coverage. The starting position is highlighted by a blue square.}
    \label{fig:heatmap_lab}
     \end{subfigure}
     \hfill
     \begin{subfigure}[b]{0.3\textwidth}
         \centering
         \includegraphics[width=\textwidth]{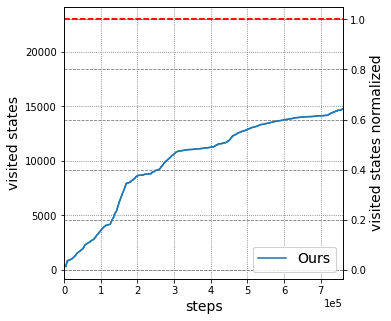}
         \label{fig: visited_state_lab}
     \end{subfigure}
     \hfill
     \caption{a) Our model heatmap coverage of the large real world with a coarse positioning of the obstacles. b) the number of visited states over steps coupled with the percentage of the environment it represents.} % {The blue curve shows the number of visited states covered over the number of total steps, the dotted red line displaying the number of states the environment possesses (23000 states)
    \label{fig:lab}
\end{figure*}

\section{Experimental Results}
\label{sec:result}

%proposition:
%results intro, what kind of results we will show and why

Our experiments are designed to highlight the agent's ability to explore a totally new environment, using only visual observations and low-level actions. 

\textbf{Environments}. The experiments are divided into two main groups. The first group is realized on Habitat \cite{habitat}, a navigation simulator showcasing a visually realistic domain. The second group studies real environments of diverse size and complexity. 
% \textbf{Agents}. 
Consistently across all environments, the tests are realized with the limited field of view of a single camera seeing in front of the agent (i.e egocentric view). Each state corresponds to a visual observation and each observation corresponds to the agent first view 64$\times$64 RGB image. The agent moves in discrete steps in the environment, each step corresponding to a motion forward (0.25m) or a turn right or left (10\textdegree). 

% 3In each test, the agent is left to explore the given environment without prior training. 

\subsection{Simulation}
\label{subsec:simulation}
% \textbf{Baseline.} 
To establish a baseline for the Habitat simulation tasks, we utilize C-BET, an exploration approach that seek out both surprising and high-impact areas of the environment, and that has been shown to perform well on this benchmark \cite{cbet}. Both our proposed model and C-BET aim to fully explore the environment by physically cover visitable states, which are represented as plan cells of 5$\times$5cm in the Habitat simulations. We conduct exploration experiments in three diverse apartments of 52,43m$^2$ averaged surface, as presented in Appendix \ref{app:simu_envs}. Results show that our agent outperforms C-BET in terms of exploration coverage and efficiency, as illustrated in Fig. \ref{fig:three appartments} for individual apartments (\ref{fig:apartment 0},\ref{fig:apartment 1},\ref{fig:apartment 2}), as well as the overall mean results across all environments (\ref{fig:all_apps}).

%//explanation why does dreamer performs better, to complete//

%SHOULD THOSE BE MENTIONED? : 
%DreamerV2 needs to output a discrete which is accomplished by using a one-hot categorical distribution. Next to this, the actor model will use REINFORCE gradients instead of straight-through gradients to update the actor model [39]. REINFORCE gradients will use the data from a complete episode to update the policy parameters. Since this is done after the episode, the agent will use the same policy during the whole episode which makes it an off-policy algorithm. Each episode will last for 500 steps and after each episode, the agent will be reset to its initial position and orientation.

\subsection{Real world}
\label{subsec:real}

\textbf{Setup.} In our real-world experiments, we employ a Kobuki TurtleBot 2i as the platform, with a Jetson AGX Orin serving as the core system. Data from the robot is sent to a computer for training after each episode, and an Intel Realsense LiDAR Camera L515 is utilized to perceive RGB and depth information, with depth data specifically used for obstacle detection and avoidance, cancelling any forward motion leading to a collision. The Robot Operating System (ROS), a framework to help build and control robots, providing tools and libraries for tasks such as communication, hardware control, and navigation is implemented as the baseline framework for executing the agent's behavior. In this context, a visitable state in the real world corresponds to a plan cell with dimensions of 10$\times$10cm.

To assess the system's performance under different conditions, we conducted two tests: one in a small, well-controlled environment, and another in a larger environment with variable conditions and potential changes happening.

\textbf{Small real-world.}  The first test composed of 2 runs is conducted in a controlled environment measuring approximately $2\times4$ square meters. Obstacles are deliberately placed in the environment to introduce complexity and challenges for navigation. It is important to note that during this test, no modifications happen in the environment (lighting or object motion), ensuring a consistent and stable environment throughout the experiment.
The agent is able to successfully autonomously navigate and return to its charging station, regardless of the room configuration. In 140k steps, it can achieve a coverage rate of over 80\% of the entire environment as displayed in the first run (Fig.\ref{fig:run 1}). The agent follows a pattern of exploration, spending approximately 5 minutes (500 steps) exploring before returning to its docking station. This pattern results in the area around the blue dot in the heat-map (Figure \ref{fig:heatmap_garage}) being more thoroughly explored compared to the farther areas of the map.
%The second run shows the stochasticity of the system has the agent got satisfied of a 60\% exploration before realising there was additional terrain to explore. 

%Several runs were executed Fig.\label{fig:run 1} and Fig.\label{fig:run 2} shows that the agent covers much of the area in average $\sim425k$ steps, which is much slower than in simulation given the area to explore. This is due to the agent automatically coming back to charge after 5min of exploration, resulting in the coverage heat-map Fig.\ref{fig:heatmap_garage} where the area around the station is much more explored than the furthest area. 

\textbf{Large real-world.}  The second environment is significantly more challenging, presenting an environment 4x times larger than in simulation and additional challenges, such as lightning changes and dynamic settings.
The large real-world environment covers an area of $10\times23$ square meters, and presents a more complex environment with various types of obstacles such as similar looking shelves, tables, chairs, and cardboards (see pictures in Appendix \ref{app:real_envs}). The lightning is susceptible to change and some obstacles got moved during the exploration phase, adding a significant layer of complexity. 

The agent can operate for up to 10 minutes before returning to its charging station due to battery limitations.
With this constraint of back and forth, the agent has difficulty exploring the full environment as in simulation. In simulation the agent just resets back to the starting point, in the real world the agent needs to physically return to it, which is to be considered while managing battery level. Fig.\ref{fig:lab} displays the environment's coverage of the agent over time and the areas more often visited. The agent starts at the blue square in the heatmap (Fig.\ref{fig:heatmap_lab}) and explores thoroughly the vicinity, missing the rest of the environment, reaching a 60\% coverage in 600k steps. 

\subsection{Qualitative results}
\label{subsec:analysis}

\begin{figure}[t!]
    \centering
     \begin{subfigure}[b]{\columnwidth}
        \includegraphics[width=8cm]{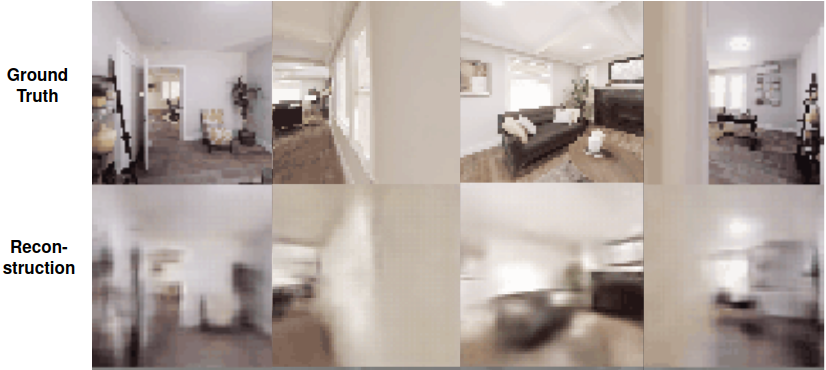}
        \caption{Simulations.} %: Randomly selected agent's observations and their predictions after 2M steps. Top row displays the ground truth, the real observations, bottom row shows the predictions made by the agent.}
        \label{fig:app0_pred_ob}
    \end{subfigure}
     \begin{subfigure}[b]{\columnwidth}
        \includegraphics[width=8cm]{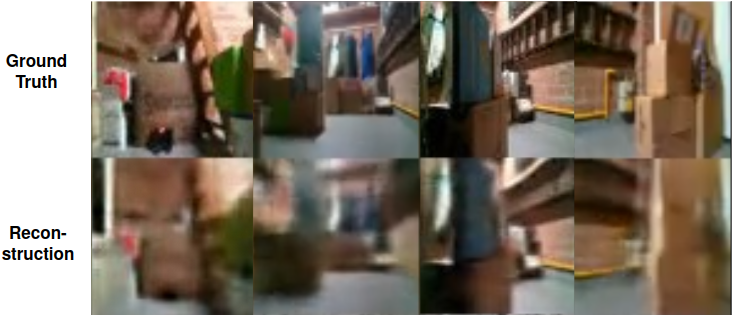}
        \caption{Small real-world.} %: Randomly selected agent's observations and their predictions. Top row shows the ground truth, the real observations, bottom row displays the corresponding predictions made by the agent.}
        \label{fig:garage_pred_ob}
    \end{subfigure}
     \begin{subfigure}[b]{\columnwidth}
        \includegraphics[width=8cm]{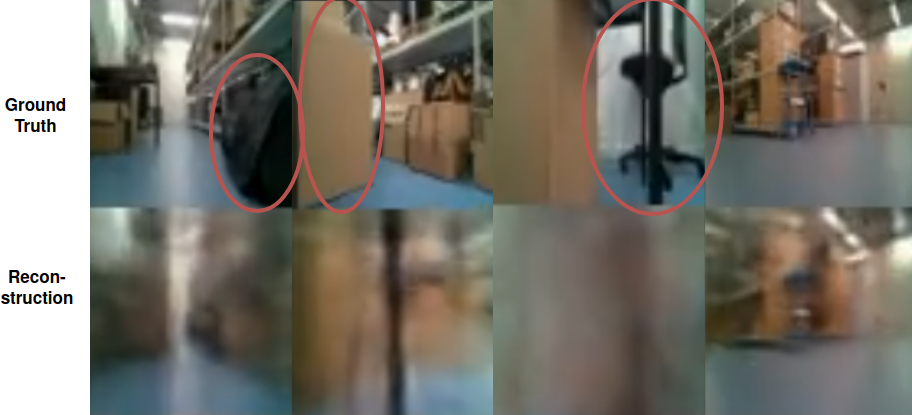}
        \caption{Large real-world.} %: Observations predictions of the agent. Top row are the ground truth, the real observations, bottom row display the predictions made by the agent. We can see that the second column shows a big difference between the ground truth and the prediction. The discrepencies, corresponding to moved objects are circled in red, those objects are not expected by the agent (cardboard and chair).}
        \label{fig:lab_pred_ob}
    \end{subfigure}
    \caption{Randomly selected agent's observations and their predictions. Top row shows the ground truth, the real observations, bottom row displays the corresponding predictions made by the agent. (c) The discrepancies, corresponding to moved objects are circled in red, those objects (i.e. (cardboard and chair)) are not expected by the agent.}
    \label{fig:pred_ob_all}
\end{figure}

In this section, we delve into the qualitative analysis of our experiments, focusing on the predicted observations generated by the world model. This adds insights into the model's ability to simulate and understand its environment and helps identifying the current limitations of our system. 
% This qualitative evaluation provides a deeper understanding of how our approach captures and represents the dynamics of the world, showing the model's capacity and limits to navigate and make informed decisions within simulated and real-world contexts.
In Figure \ref{fig:pred_ob_all}, we present the agent's observation predictions (bottom rows) alongside the real observations (top rows), displaying the agent ability to correctly infer the observation a motion should produce. 

For the simulations and small real-world experiments, Figure \ref{fig:app0_pred_ob} and \ref{fig:garage_pred_ob} show that at the end of its training, our agent is able to correctly predict observations. This shows the agent is able to learn a coherent world model of the environment and it explains why the exploration process was efficient and successful. % use it to navigate efficiently.

In the large real-world environment (Fig.\ref{fig:lab_pred_ob}), the predictions are not always so correctly inferred. When facing changes in the environment, the agent generates different expectations about what is to be seen considering its position, erroneously perceiving the dynamic state as a new area to explore for the agent. This can be observed in the red-circled areas in the Figure, showing that the agent's predictions miss the object present in the ground truth images (showing objects that were moved during the exploration process). This shows a major limitation of the current model, not able to adapt for accounting for those object's motions and consistently predicting erroneous object positions after changes in the environment. 

%\section{Analysis}
%I wouldn't have a full analysis section for a workshop. what would you add there that has not already been covered?

%===============================================================================

\section{Conclusion}
\label{sec:conclusion}

This work presents a system that merges curiosity-driven exploration with world models to achieve autonomous navigation in both simulated and real-world environments. Our study highlights the challenges of scaling autonomous navigation and exploration, progressing from an analysis in simulation to real-world dynamic and uncontrolled environments. The integration of curiosity-driven exploration and world models aims to enhance the adaptability and autonomy of robotic agents. The experimental results demonstrate improved coverage and exploration speed compared within the Habitat simulation. 

Transitioning to real-world experiments, the model performs well in a controlled "small-scale" environment despite hardware limitations. However, exploration in a larger uncontrolled environment reveals the challenges associated with scalability and real-world complexities, including changing lighting conditions and a dynamic environment. While the agent exhibits effective exploration in its vicinity, scalability and adaptability challenges become apparent, with the agent largely revisiting the same areas, and the world model failing to update object positions after displacement.

Our work emphasizes the importance of scalability, robustness, and real-world experimentation in evaluating navigation systems for practical scenarios. While our approach holds promise in controlled settings, addressing the complexities of larger and dynamic environments remains a crucial step toward achieving practical autonomy. The combined approach of world models, curiosity-driven exploration, and reinforcement learning shows potential for enabling autonomous navigation and exploration in diverse and challenging environments. Nonetheless, there are limitations to address, such as the adaptability of the world model, the diminishing efficacy of curiosity-driven exploration in complex and dynamic environments, and the need for adaptation strategies when performing lifelong learning in large-scale environments, which require substantial data for learning about the environment.

% In summary, our findings underscore both the challenges and prospects of applying these techniques to real-world scenarios. To further advance this particular study, enhancing the scalability and adaptability of world models combined with reinforcement learning, conducting additional tests in progressively larger and more intricate real-world environments, and benchmarking against various state-of-the-art models are avenues worth pursuing. 

\section*{Acknowledgment}
This research received funding from the Flemish Government (AI Research Program). 
Pietro Mazzaglia is funded by a Ph.D. grant of the Flanders Research Foundation (FWO).

%================================================================================

% \clearpage
% \newpage 

% \bibliographystyle{unsrtnat}
\bibliographystyle{IEEEtran}
\bibliography{IEEEabrv,main}

\clearpage
\onecolumn
\appendix
% \section{Environments}

\subsection{Model hyperparameters}
\label{app:hyperparameters}

Here, we report the hyperparameters for the world model, actor and critic networks, following the experimental setting of \cite{Hafner2022DreamerV2}. 

\begin{table}[h!]
\centering
\begin{tabular}{lc}
\toprule
\textbf{Name} & \textbf{Value} \\
\midrule
World Model \\
\midrule
Batch size & 50 \\
Batch sequence length & 50 \\
Discrete latent state dimension & 32 \\
Discrete latent classes & 32 \\
GRU cell dimension & 200 \\
KL free nats & 1 \\
KL balancing & 0.8 \\
Adam learning rate & $3\cdot10^{-4}$ \\
Slow critic update interval & 100 \\
\midrule
Actor-Critic \\
\midrule
Imagination horizon & 15 \\
$\gamma$ parameter & 0.99 \\
$\lambda$ parameter  & 0.95 \\
Adam learning rate & $8\cdot10^{-5}$ \\
Actor entropy loss scale & $1\cdot10^{-4}$ \\
\midrule
Common \\
\midrule
MLP number of layers & 4 \\
MLP number of units & 400 \\
Hidden layers dimension & 400 \\
Adam epsilon & $1\cdot10^{-5}$ \\
Weight decay & $1\cdot10^{-6}$ \\
Gradient clipping & 100 \\
\bottomrule
\end{tabular}
\vspace{1em}
\caption{World model, actor-critic and common hyperparameters.}
\label{tab:hparams}
\end{table}

\subsection{Simulation}
\label{app:simu_envs}
The simulated environments, namely apartment 0, apartment 1, and apartment 2, were sourced from the Replica-Dataset \cite{replica_dataset}. This dataset contains photo-realistic scenes used in our experiments.

\begin{figure}[htb!]
     \centering
     \begin{subfigure}[b]{0.3\textwidth}
         %\centering
    \includegraphics[width=\textwidth]{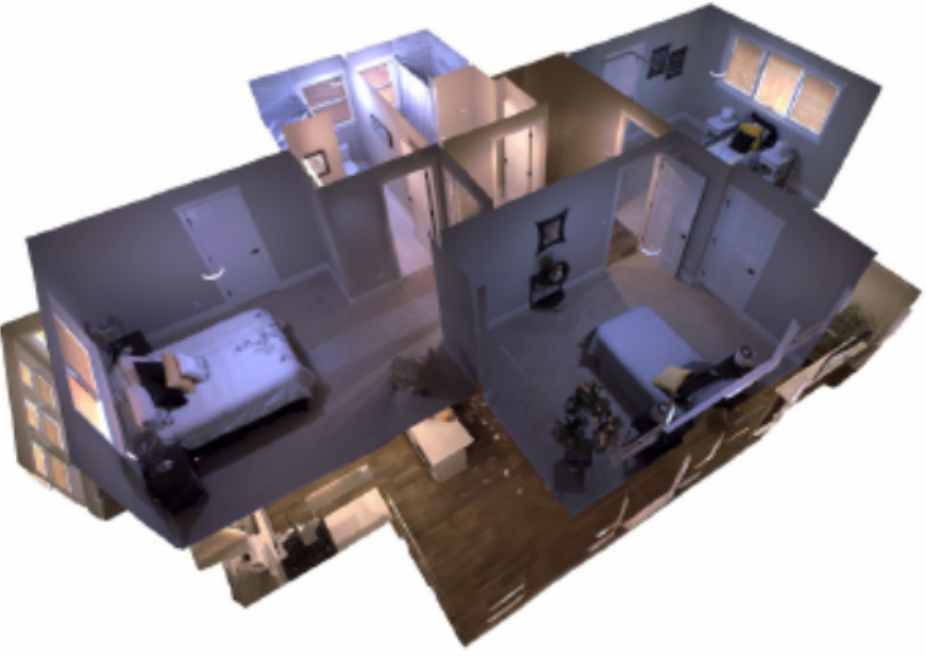}
    \caption{Apartment 0}
    \label{fig:app0}
     \end{subfigure}
     \hfill
     \begin{subfigure}[b]{0.3\textwidth}
         %\centering
         \includegraphics[width=\textwidth]{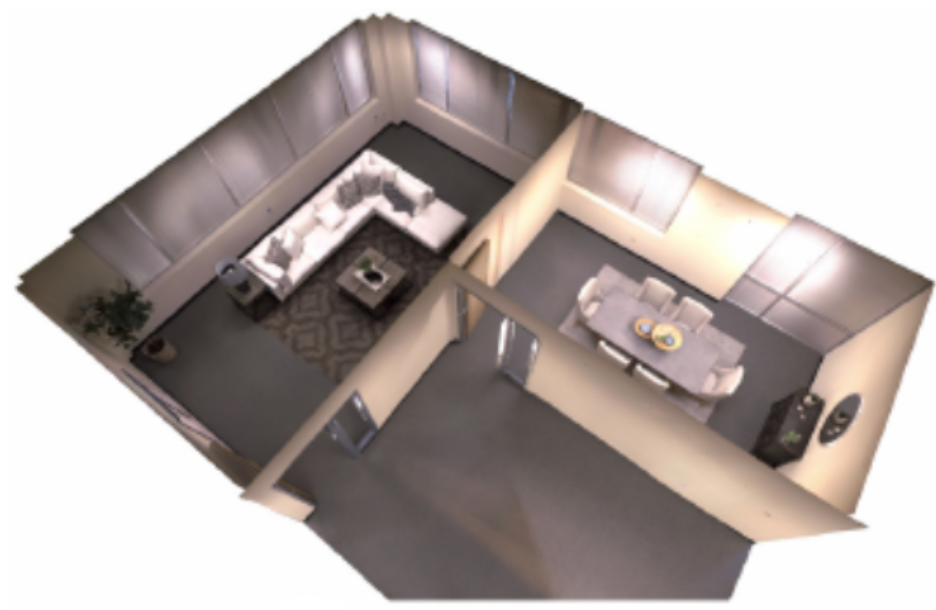}
         \caption{Apartment 1}
         \label{fig:app 1}
     \end{subfigure}
     \begin{subfigure}[b]{0.3\textwidth}
         %\centering
         \includegraphics[width=\textwidth]{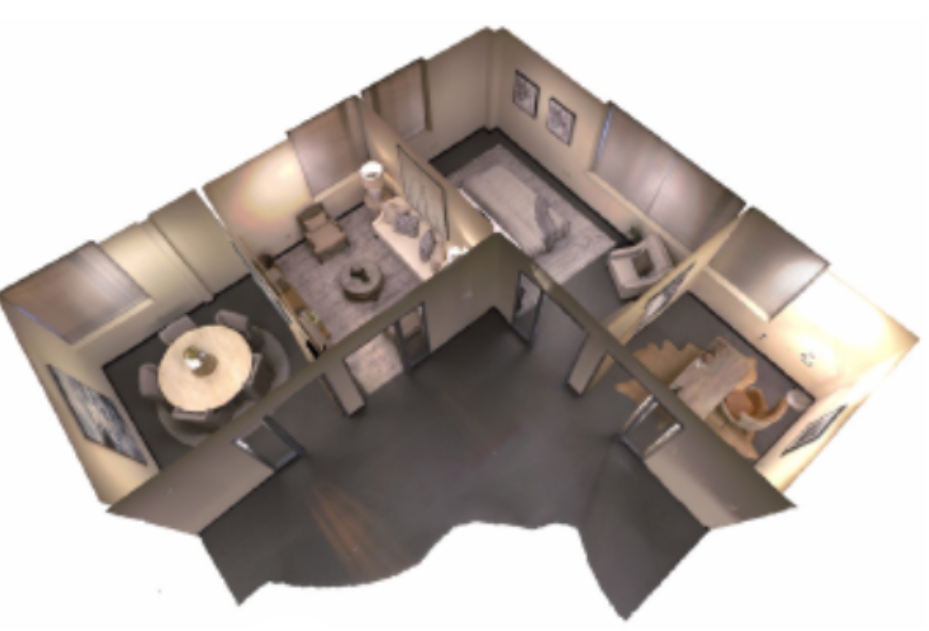}
         \caption{Apartment 2}
         \label{fig:app 2}
     \end{subfigure}
     \hfill
        \caption{Habitat simulation environments extracted from the Replica-Dataset}
        \label{fig:three appartments appendix}
\end{figure}

Those apartments are designed as real apartments, proposing large spaces to visit as is shown in Tab.\ref{tab:apartement_size}.
\begin{table}[!htb]
\centering
\begin{tabular}{|c|c|c|c|}
\hline
 & apartement 0 & apartement 1 & apartement 2 \\ \hline
surface & $58.69 m^2$ & $53.32 m^2$ & $45.31 m^2$ \\ \hline
\begin{tabular}[c]{@{}c@{}}visitable \\ states\end{tabular} & 23476 & 21328 & 18123 \\ \hline
\end{tabular}
\vspace{+2mm}
\caption{Apartments number of visitable states (cells of 5 by 5 cm) and equivalent surface in $m^2$.}
\label{tab:apartement_size}
\end{table}

C-BET, for which we adopt the open-source implementation (available at \url{https://github.com/sparisi/cbet}), underwent a training process following the protocol outlined in the study by \cite{cbet}, involving 5 million training steps. A comparative analysis of the final exploration coverage after full training is presented in Fig.\ref{fig:apps_comparison}, revealing that our model, trained for 2 million steps, achieved commendable coverage. Despite the extended training duration of 5 million steps, C-BET exhibited significantly lower exploration coverage in apartments 0 and 1. Apartment 2 exhibited comparable exploration between the two models. Notably, our model demonstrates a more efficient exploration strategy, while C-BET exhibited a tendency to revisit certain areas before venturing into new regions.
 
\begin{figure}[!htb]
    \centering
    \includegraphics[width=8cm]{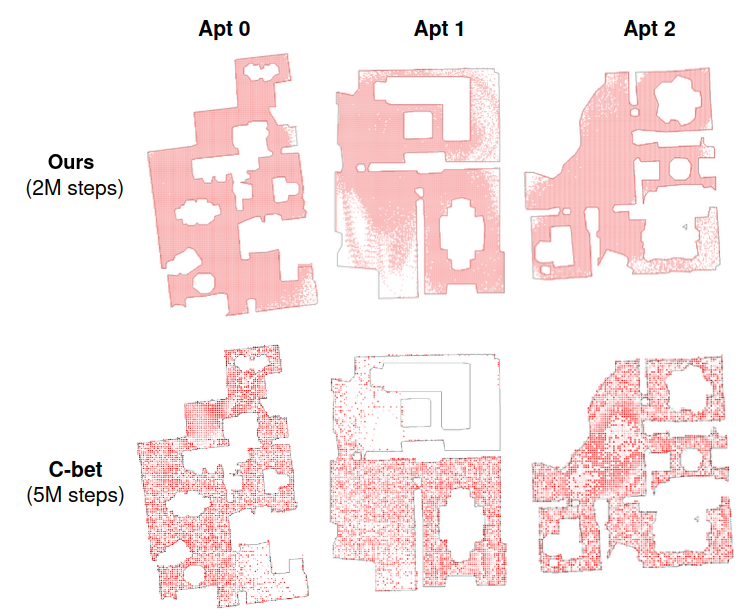}
    \caption{Top view of the best heat-map coverage of the 3 apartments by ours and C-BET model after their attributed maximum number of steps. }
    \label{fig:apps_comparison}
    \vspace{-0mm}
\end{figure}

\subsection{Real world}
\label{app:real_envs}
The following section presents images providing visual insight into the real-world environments utilized for testing purposes. These pictures showcase the spaces where our navigation techniques were evaluated.

\begin{figure}[!htb]
    \centering
    \includegraphics[width=9cm]{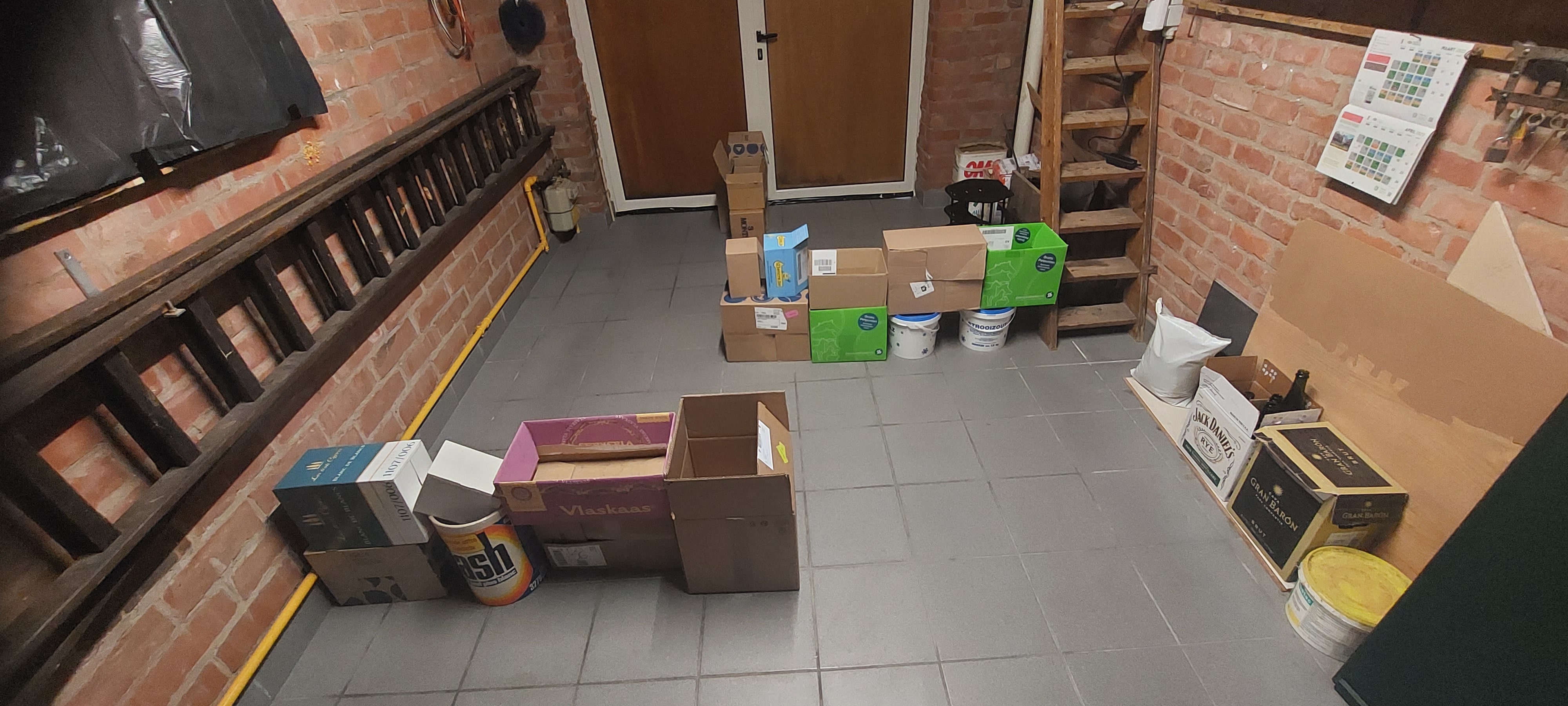}
    \caption{Picture of the setup of the first real-world environment :"Small Real World".}
    \label{fig:garage_pic}
    % \vspace{-20mm}
\end{figure}

\begin{figure}[htb!]
     \centering
     \begin{subfigure}[b]{0.8\textwidth}
         %\centering
    \includegraphics[width=\textwidth]{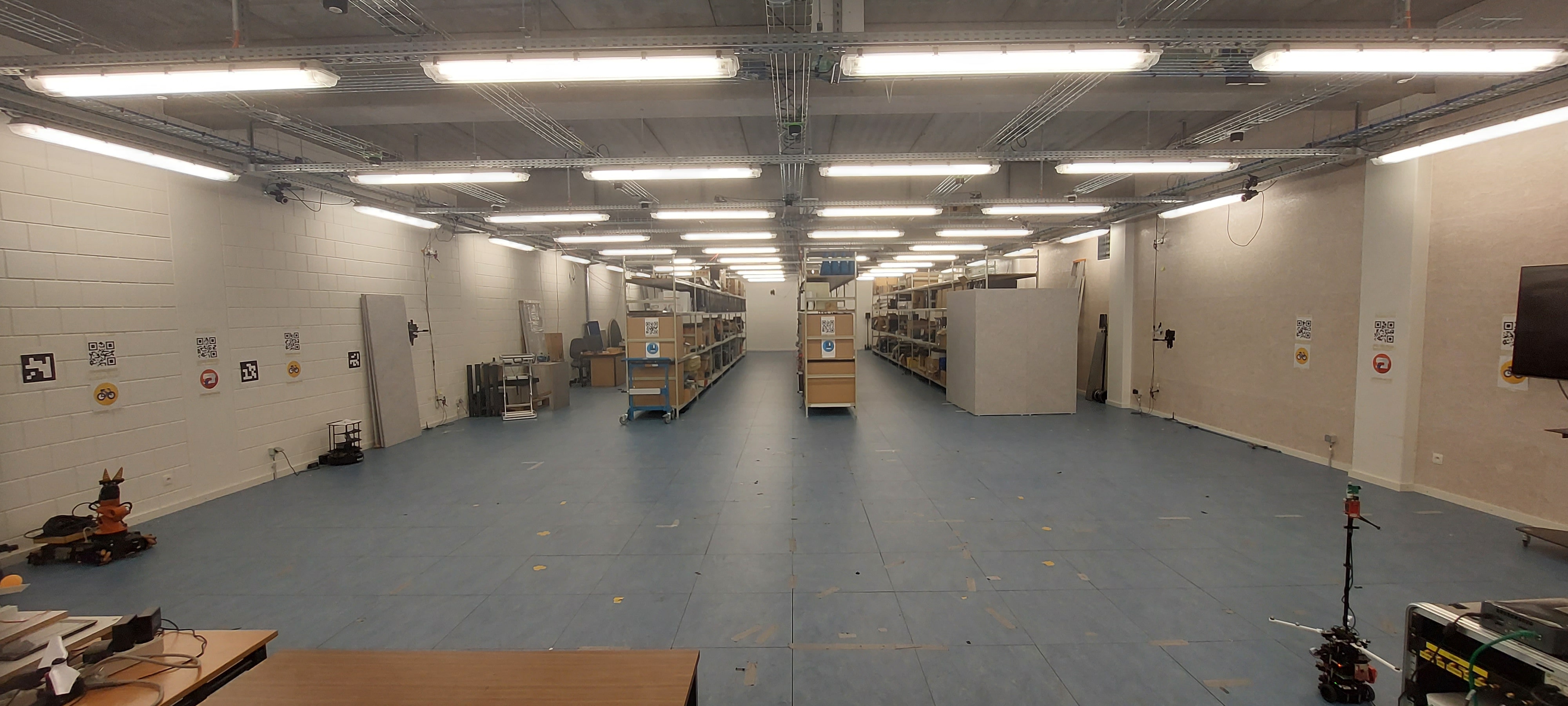}
    \caption{lab overall landscape}
    \label{fig:iiot_landscape}
     \end{subfigure}
     \hfill
     \begin{subfigure}[r]{0.23\textwidth}
         %\centering
         \includegraphics[width=\textwidth]{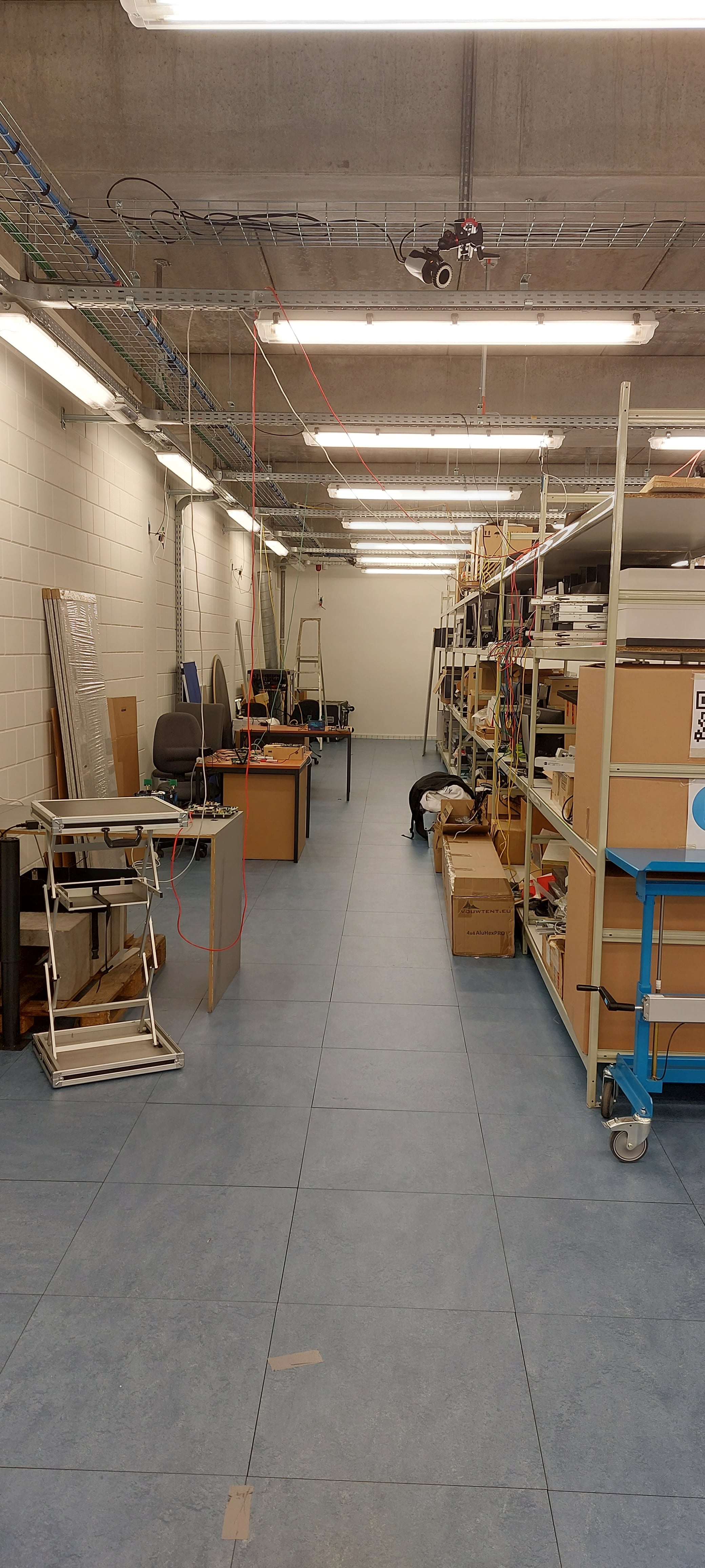}
         \caption{utter left aisle of the lab}
         \label{fig:iiot_aisle1}
     \end{subfigure}
     \begin{subfigure}[r]{0.23\textwidth}
         %\centering
         \includegraphics[width=\textwidth]{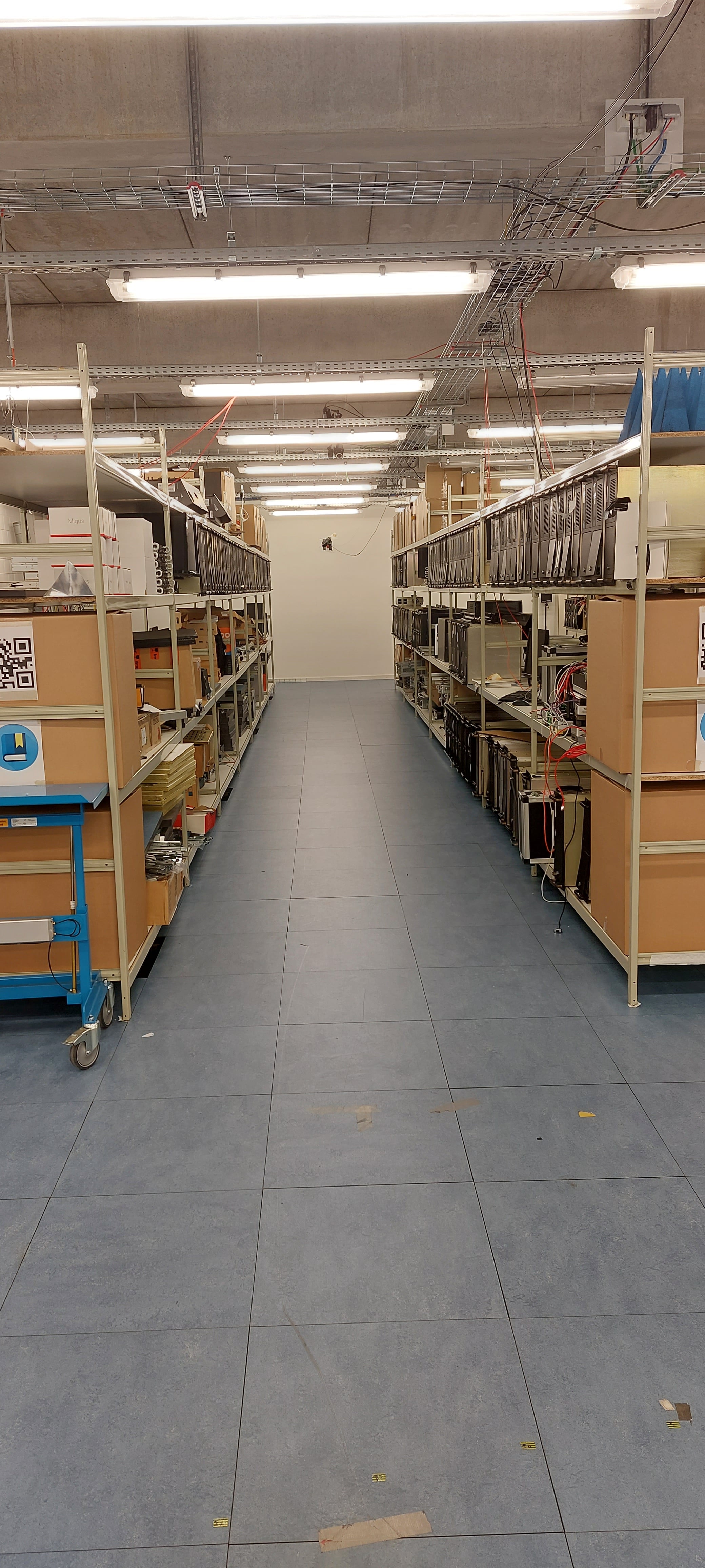}
         \caption{center left aisle of the lab}
         \label{fig:iiot_aisle2}
     \end{subfigure}
     \begin{subfigure}[r]{0.23\textwidth}
         %\centering
         \includegraphics[width=\textwidth]{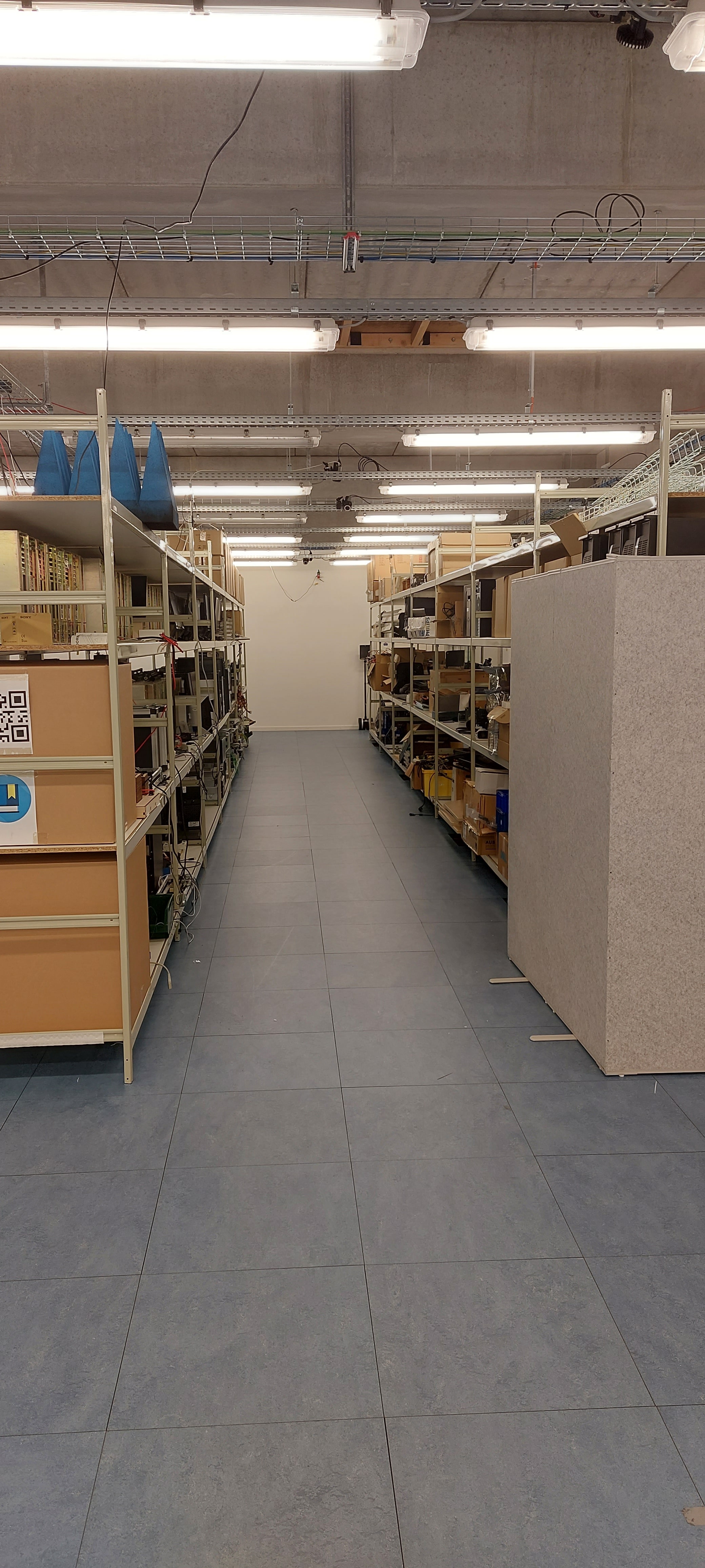}
         \caption{center right aisle of the lab}
         \label{fig:iiot_aisle3}
     \end{subfigure}
     \begin{subfigure}[r]{0.23\textwidth}
         %\centering
         \includegraphics[width=\textwidth]{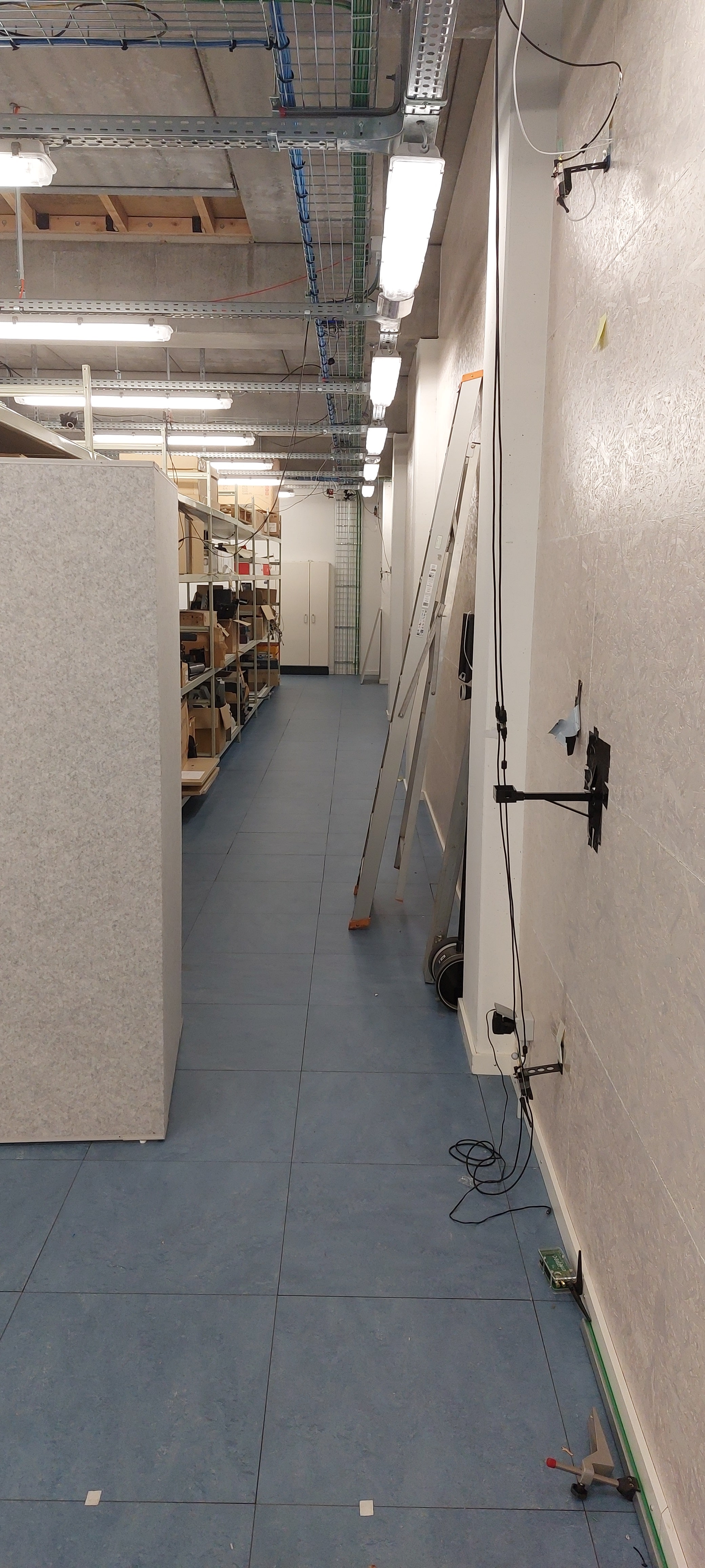}
         \caption{utter right aisle of the lab}
         \label{fig:iiot_aisle4}
    \end{subfigure}
     \hfill
        \caption{Picture of the initial setup of the large real-world environment.}
        \label{fig:iiot_lab_pics}
\end{figure}

\clearpage

\subsection{Auto docking}
\label{app:dock}
The agent is able to determine its own path to reach the docking station upon completion of its exploration or depletion of its battery. However, sometimes it failed and thus required human intervention to plug it back into its station manually.
The main cause of not getting to the docking station is due to too much drift on the wheels of the robot, leading to a very inaccurate position estimation. In a small environment, the agent corrects its position upon docking; however, in a large environment, it leads to failure as it is not able to locate the docking station that is localized by the agent up to 1m from it.

\begin{table}[htb!]
\centering
\begin{tabular}{|c|c|c|c|}
\hline
 & requests & success & success ratio \\ \hline
\begin{tabular}[c]{@{}c@{}}Small world\\ auto docking\end{tabular} & 596 & 572 & 96\% \\ \hline
\begin{tabular}[c]{@{}c@{}}Large world\\ auto docking\end{tabular} & 1023 & 758 & 74\% \\ \hline
\end{tabular}
\caption{Number of auto-docking request and success ratio in each environment.}
\end{table}

\begin{figure}[!htb]
    \centering
    \includegraphics[width=0.5\textwidth]{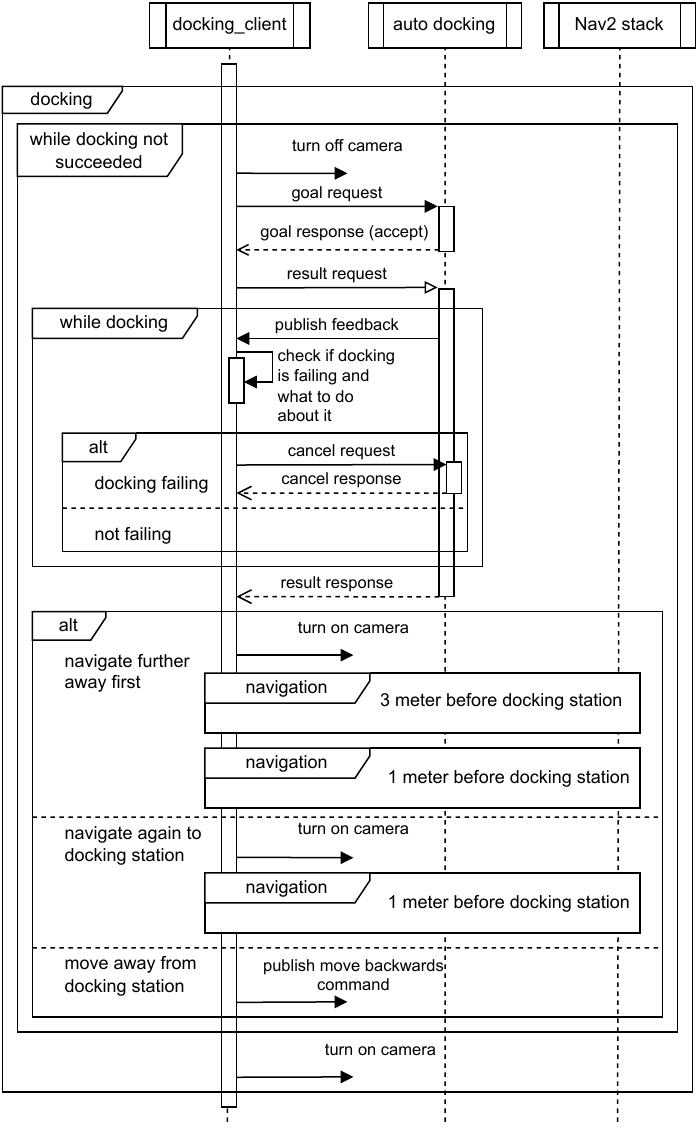}
    \caption{The docking client node requests a docking goal to the auto docking action server. Recovery systems are used to make sure the docking process succeeds. }
    \label{fig:apps_comparison}
    \vspace{-0mm}
\end{figure}

\end{document}